\documentclass[10pt,twocolumn,letterpaper]{article}

\pdfoutput=1
\usepackage{cvpr}
\usepackage{times}
\usepackage{epsfig}
\usepackage{graphicx}
\usepackage{amsmath}
\usepackage{amssymb}
\usepackage{multirow}
\usepackage{subfigure}

\usepackage{array}
\newcolumntype{L}[1]{>{\raggedright\let\newline\\\arraybackslash\hspace{0pt}}m{#1}}
\newcolumntype{C}[1]{>{\centering\let\newline\\\arraybackslash\hspace{0pt}}m{#1}}
\newcolumntype{R}[1]{>{\raggedleft\let\newline\\\arraybackslash\hspace{0pt}}m{#1}}

\usepackage[breaklinks=true,bookmarks=false]{hyperref}

\cvprfinalcopy % *** Uncomment this line for the final submission

\def\cvprPaperID{1730} % *** Enter the CVPR Paper ID here

\begin{document}

%%%%%%%%% TITLE
\title{Creating Something from Nothing:\\Unsupervised Knowledge Distillation for Cross-Modal Hashing}

\author{Hengtong Hu\textsuperscript{1,2}\thanks{This work was done when the first author was an intern at Huawei Noah's Ark Lab.},\quad Lingxi Xie\textsuperscript{3},\quad Richang Hong\textsuperscript{1,2}\thanks{Corresponding author.},\quad Qi Tian\textsuperscript{3}\\
\textsuperscript{1}School of Computer Science and Information Engineering, Hefei University of Technology, \\%,\quad
\textsuperscript{2}Key Laboratory of Knowledge Engineering with Big Data, Hefei University of Technology, \\%,\quad
\textsuperscript{3}Huawei Inc.\\
{\tt\small huhengtong.hfut@gmail.com},\quad{\tt\small 198808xc@gmail.com},\quad{\tt\small hongrc@hfut.edu.cn},\quad{\tt\small tian.qi1@huawei.com}
}

\maketitle
%\thispagestyle{empty}

%%%%%%%%% ABSTRACT
\begin{abstract}
%\textcolor{blue}{
In recent years, cross-modal hashing (CMH) has attracted increasing attentions, mainly because its potential ability of mapping contents from different modalities, especially in vision and language, into the same space, so that it becomes efficient in cross-modal data retrieval. There are two main frameworks for CMH, differing from each other in whether semantic supervision is required. Compared to the unsupervised methods, the supervised methods often enjoy more accurate results, but require much heavier labors in data annotation. In this paper, we propose a novel approach that enables guiding a supervised method using outputs produced by an unsupervised method. Specifically, we make use of teacher-student optimization for propagating knowledge. Experiments are performed on two popular CMH benchmarks, i.e., the MIRFlickr and NUS-WIDE datasets. Our approach outperforms all existing unsupervised methods by a large margin. %It even gets close to the accuracy level of supervised learning but remains completely annotation-free.}
\end{abstract}

%%%%%%%%% BODY TEXT
\section{Introduction}

Recently, with the rapid increase of multimedia data, cross-modal retrieval~\cite{wu2017online,zhang2016collaborative,kiros2014unifying,zhang2018attention,deng2018triplet,bronstein2010data,feng2014cross,liu2012supervised,lin2015semantics} has attracted more and more attentions in both academia and industry. The goal is to retrieve instances from one modality using a query instance from another modality, \textit{e.g.}, finding an image with a few textual tags. One of the most popular pipeline for this purpose, named cross-modal hashing (CMH)~\cite{bronstein2010data,kumar2011learning,liu2017cross,zhang2018attention,deng2018triplet}, involves mapping contents in different modalities into a common hamming space. By compressing each instance into a fixed-length binary code, the storage cost can be dramatically reduced and the time complexity for retrieval is constant since the indexing structure is built with hashing codes.
%and, in order to reduce the storage and retrieval costs in a large-scale system, compressing each instance into a fixed-length binary code.

%In recent years cross-modal retrieval has attracted more and more attention owing to the explosive increase in multimedia data from a great variety of search engines and social media[……..]. Cross-modal retrieval has rich application scenarios in our life. For example, we usually want to search some interesting photos through several simple tags like place and time. In order to process large-scale multi-modal data, hashing based methods[…….] which aim to map multi-modal data into binary codes have been proposed. By using the binary codes, the storage cost and computation cost can greatly decrease. 

State-of-the-art CMH methods can be roughly categorized into two parts, namely, supervised and unsupervised methods. Both of them learn to shrink the gap between the distributions of two sets of training data (\textit{e.g.}, using adversarial-learning-based approaches~\cite{li2018self,li2019coupled,gu2018look}), but they differ from each other in whether an instance-level annotation is provided during the training stage. From this perspective, the supervised CMH methods~\cite{bronstein2010data,liu2012supervised,lin2015semantics,wang2013learning,zhang2014large}, receiving additional supervision, often produce more accurate results, and the unsupervised counterparts, while achieving lower performance, are relatively easier to deploy to real-world scenarios.

%Cross-modal hashing can be generally categorized into supervised methods and unsupervised methods by whether requiring label information. Supervised cross-modal hashing methods[……] aim to sufficiently explore semantic relevance by labels to bridge the modality gap. Superior performance can be achieved by exploiting abundant semantic information for supervised cross-modal hashing. However, they always require amounts of manually annotated labels which make them infeasible in real-world applications. Unsupervised cross-modal hashing methods usually learn hashing functions by preserving the original relationship among unlabeled cross-modal data. Mostly they exploit correlation information according to the pairwise relationship naturally exists in cross-modal data. Without using annotation information, unsupervised cross-modal hashing is flexible to apply in real world. Meanwhile, lacking semantic information results in that unsupervised hashing methods always achieve uncompetitive performance. 

% \begin{figure}
% \centering
% %\fbox{\rule{0pt}{2in} \rule{0.9\linewidth}{0pt}}
% \includegraphics[width=0.9\linewidth]{illustration.pdf}
% \caption{\textcolor{red}{The illustration of our idea, unsupervised knowledge distillation for unsupervised cross-modal hashing. We guide a supervised method using the outputs produced by an unsupervised method to improve unsupervised cross-modal hashing.}}
% \label{fig:illustration}
% \end{figure}
\begin{table}[t]
%\centering
\begin{center}
\begin{tabular}{|l|c|c|c|}
\hline
      & WL? & ER? & KD?   \\ \hline \hline
DCMH~\cite{jiang2017deep}  &     & \checkmark    &           \\ \hline
SSAH~\cite{li2018self}  &     & \checkmark    &              \\ \hline
UCH~\cite{li2019coupled}   & \checkmark    &     &              \\ \hline
UGACH~\cite{zhang2018unsupervised} & \checkmark    & \checkmark    &       \\ \hline
UKD  & \checkmark    & \checkmark     & \checkmark         \\ \hline
\end{tabular}
\end{center}
\caption{The difference between our approach and some recent cross-modal hashing methods. Here, `WL' indicates that training without using labels, `ER' indicates that the method utilizes  extensive relevance information rather than only the pairwise information, and `KD' indicates utilizing knowledge distillation in the training process.}
\label{attributes}
\end{table}

This paper combines the benefits of both methods by a simple yet effective idea, known as \textbf{creating something from nothing}. The core idea is straightforward: the supervised methods do not \textit{really} require each instance to be labeled, but they use the labels to estimate the similarity between each pair of cross-modal data. Such information, in case of no supervision, can also be obtained from calculating the distance between their feature vectors, with the features provided by a trained unsupervised CMH method. Our approach, \textbf{unsupervised knowledge distillation} (UKD), contains an unsupervised CMH module followed by another supervised one, both of which can be freely replaced by new and more powerful models in the future.

%Therefore, how to effectively bridge the modality gap without using manual annotations becomes crucial to achieve satisfactory performance for cross-modal hashing. Supervised cross-modal hashing methods usually utilize labels to acquire similarity information of cross-modal data, for example, an image is relevant to a text if they have similar labels. Noticeably, an unsupervised model just aim to output relevant instances for queries, which can be used to provide this similarity information. Imagining such a situation where an unsupervised model combines with a supervised model, the former firstly learns reliable similarity information and then guides the optimization of the later. Then without any label information, more reliable hashing functions can be learned by the powerful learning ability of supervised models. The key is to find an effective way to combine supervised and unsupervised cross-modal hashing models.

Our research paves the way towards an interesting direction that using an unsupervised method to guide a supervised method, for which CMH is a good scenario to test on. We perform experiments on two popular cross-modality retrieval datasets, \textit{i.e.}, MIRFlickr and NUS-WIDE, and demonstrate state-of-the-art performance, outperforming existing unsupervised CMH methods by a significant margin. Moreover, we delve deep into the benefits of supervision, and point out a few directions for future research.

The remainder of this paper is organized as follows. Section~\ref{related_work} briefly reviews the preliminaries of cross-modal retrieval and hashing, and Section~\ref{approach} describes the unsupervised knowledge distillation approach. Experimental results are shown in Section~\ref{experiments} and conclusions are drawn in Section~\ref{conclusions}.

%The contributions of this work can be summarized as follows. First, we propose a novel approach that enables guiding a supervised cross-modal hashing method using outputs produced by an unsupervised method. Second, we achieve this by making use of teacher-student optimization. To the best of our knowledge, we are the first to formulate a distillation method for cross-modal hashing. Third, extensive experiments on two benchmark datasets demonstrate that our approach outperforms existing start-of-the-art performances unsupervised cross-modal hashing methods and even gets close to the accuracy level of supervised methods. 

%-------------------------------------------------------------------------
\section{Related Work}
\label{related_work}

\subsection{Cross-Modal Retrieval and Hashing}

Cross-modal retrieval aims to search semantically similar instances in one modality using a query from another modality~\cite{wu2017online,xu2019deep}. Throughout this paper, we consider the retrieval task between vision and language, \textit{i.e.}, involving images and texts. To map them into the same space, two models need to be trained, one for each modality. The goal is to make the image-text pairs with relevant semantics to be close in the feature space. To train and evaluate the mapping functions, a dataset with image-text pairs is present. The dataset is further split into a training set and a query set, \textit{i.e.}, the testing stage is performed on the query set. %Some examples of paired data are shown in Figure~\ref{fig:problem}.

In the past decade, many efforts were made on this topic~\cite{kiros2014unifying,zhang2016collaborative,wu2017online}. However, most of them suffered from high computation costs in real-world, high-dimensional data. To scale up these models to real-world scenarios, researchers often compressed the output of these models into binary vectors of a fixed length~\cite{bronstein2010data,kumar2011learning,feng2014cross,liu2014discrete}, \textit{i.e.}, Hashing codes. In this situation, this task is often referred to as cross-modality hashing.

\subsection{Supervised Cross-Modal Hashing Methods}

The fundamental challenges of cross-modal hashing lie in learning reliable mapping functions to bridge the modality gap. Supervised methods~\cite{liu2012supervised,zhang2018attention,wu2018learning,li2018self,xu2019deep,deng2018triplet} achieved this goal by exploiting semantic labels to capture rich correlation information among data from different modalities. Traditional supervised learning methods were mostly based on handcrafted features, and aimed to understand the semantic relevance in the common space. SePH~\cite{lin2015semantics} proposed a semantics-preserving hashing method which aimed to approximate the distribution of semantic labels with hash codes on the Hamming space via minimizing the KL-divergence. Wang~\textit{et al.}~\cite{wang2013learning} proposed to leverage list-wise supervision into a principled framework of learning the hashing function.

With the rapid development of deep learning, researchers started to build supervised methods upon more powerful yet discriminative features. DCMH~\cite{jiang2017deep} proposed a deep cross-modal hashing method by integrating feature learning and binary quantization into one framework. SSAH~\cite{li2018self} improved this work by proposing a self-supervised approach, which incorporated adversarial learning into cross-modal hashing. Zhang~\textit{et al.}~\cite{zhang2018attention} also investigated a similar idea by proposing an adversarial hashing network with an attention mechanism to enhance the measurement of content-level similarities. These supervised methods achieved superior performance, arguably by acquiring correlation information from the semantic labels of both images and texts. However, acquiring a large amount of such labels is often expensive and thus intractable, which makes the supervised approaches infeasible in the real-world applications.

\subsection{Unsupervised Cross-Modal Hashing Methods}

% \begin{figure}
% \centering
% %\fbox{\rule{0pt}{2in} \rule{0.9\linewidth}{0pt}}
% \includegraphics[width=0.9\linewidth]{retrieval_problem.pdf}
% \caption{An illustration of the task of cross-modal retrieval. Given instances from one modality, it is required to return a rank-list of instances from another modality.}
% \label{fig:problem}
% \end{figure}
\begin{figure*}
\centering
\includegraphics[width=0.9\linewidth]{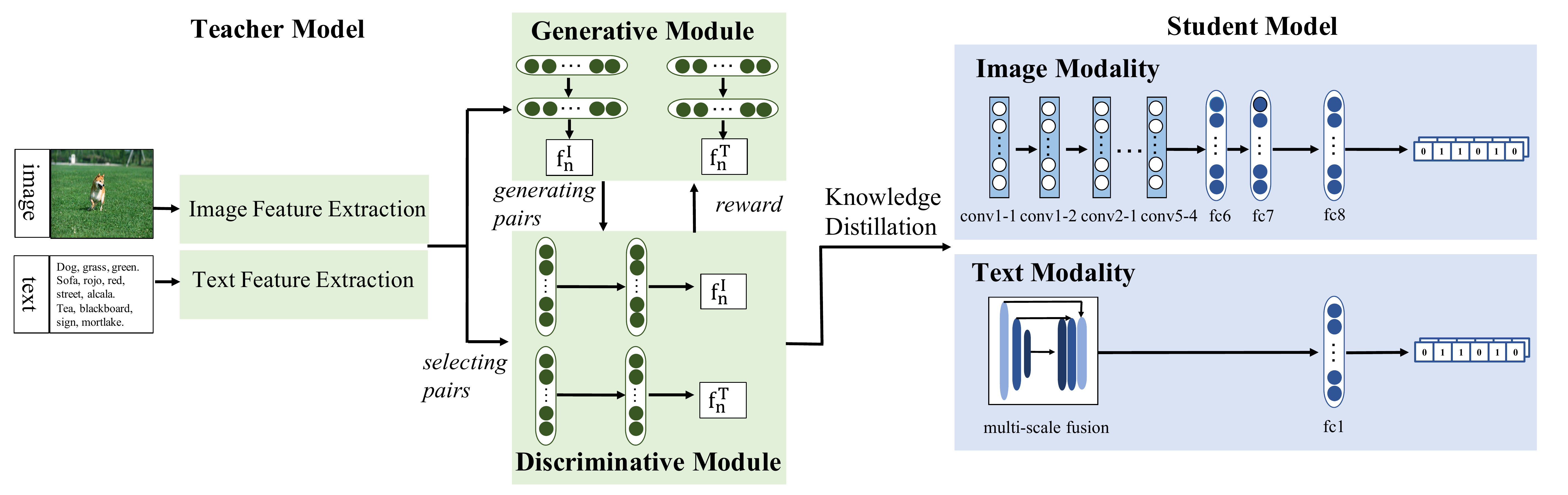}
\caption{The proposed UKD framework which involves training a teacher model in an unsupervised manner, constructing the similarity matrix $\mathbf{S}$ by distilling knowledge from the teacher model, and using $\mathbf{S}$ to supervise the student model. Each dot represents an intermediate feature. Please zoom in to see the details of this figure.}
\label{fig:framework}
\end{figure*}

Compared with the supervised counterparts, unsupervised cross-modal hashing methods~\cite{ding2014collective,zhang2018unsupervised,gu2018look,wu2018cycle} only relied on the correlation information from the paired data, making it easier to be deployed to other scenarios. These methods usually learned hashing codes by preserving inter- and intra-correlations. For example, Song~\textit{et al.}~\cite{song2013inter} proposed inter-media hashing to establish a common Hamming space by maintaining inter-media and intra-media consistency. Recently, several works introduced deep learning to improve unsupervised cross-modal hashing. UGACH~\cite{zhang2018unsupervised} utilized a generative adversarial network to exploit the underlying manifold structure of cross-modal data. As an improvement, UCH~\cite{li2019coupled} coupled the generative adversarial network to build two cycled networks in an unified framework to learn common representations and hash mapping simultaneously.

Despite the superiority in reducing the burden of data annotation, the accuracy of unsupervised cross-modal hashing methods is often below satisfaction, in particular, much lower than the supervised counterparts. The main reason lies in lacking the knowledge of pairwise similarity for the training data pairs. On the other hand, we notice that the output of an unsupervised model contains, though somewhat inaccurate, such semantic information. This motivates us to guide a supervised model by the output of an unsupervised model. This is yet another type of research which distills knowledge to assist model training.

\section{Our Approach}
\label{approach}

In this work, we focus on the idea named \textbf{creating something from nothing}, \textit{i.e.}, a supervised cross-modal hashing method can be guided by the output of an unsupervised method, which reveals the similarity between training data pairs. Figure~\ref{fig:framework} shows the framework of the proposed UKD.  In what follows, we first explain the motivation of our approach, and then introduce the proposed pipeline, \textbf{unsupervised knowledge distillation}, from two aspects, namely, how to distill similarity from an unsupervised model, and how to utilize it efficiently to optimize a supervised model.

%-------------------------------------------------------------------------

\subsection{Supervised and Unsupervised Baselines}

Throughout this paper, we consider the case that the training set contains paired data, \textit{i.e.}, $\mathcal{D}=\left\{\mathbf{v}_n^\mathrm{I},\mathbf{v}_n^\mathrm{T}\right\}_{n=1}^N$, where $N$ is the number of image-text pairs. Here, $\mathbf{v}_n^\mathrm{I}\in\mathbb{R}^{D_\mathrm{I}}$ be an image and $\mathbf{v}_n^\mathrm{T}\in\mathbb{R}^{D_\mathrm{T}}$ be a text, where the superscripts $\mathrm{I}$ and $\mathrm{T}$ denote `images' and `texts', and $D_\mathrm{I}$ and $D_\mathrm{T}$ denote the dimensionality of the feature spaces, respectively. $D_\mathrm{I}$ and $D_\mathrm{T}$ can be different, \textit{e.g.}, as in our experiments. The models that map them into the same space are denoted as $\mathbf{f}_n^\mathrm{I}\doteq f^\mathrm{I}\!\left(\mathbf{v}_n^\mathrm{I};\boldsymbol{\theta}^\mathrm{I}\right)\in\mathbb{R}^K$ and $\mathbf{f}_n^\mathrm{T}\doteq f^\mathrm{T}\!\left(\mathbf{v}_n^\mathrm{T};\boldsymbol{\theta}^\mathrm{T}\right)\in\mathbb{R}^K$, respectively, where $K$ is the dimensionality of the common feature space, and $\boldsymbol{\theta}^\mathrm{I}$ and $\boldsymbol{\theta}^\mathrm{T}$ are model parameters. The compressed hashing code for images and texts are denoted by $\mathbf{b}_n^\mathrm{I}\doteq \mathrm{sgn}\!\left(\mathbf{f}_n^\mathrm{I}\right)$ and $\mathbf{b}_n^\mathrm{T}\doteq \mathrm{sgn}\!\left(\mathbf{f}_n^\mathrm{T}\right)$, respectively, \textit{i.e.}, both $\mathbf{b}_n^\mathrm{I}$ and $\mathbf{b}_n^\mathrm{T}$ fall within $\left\{-1,+1\right\}^K$.

The key to cross-modal hashing lies in recognizing which pairs of image-text data are semantically relevant while other are not, so that the model can learn to pull the features of relevant pairs closer in the common space. A straightforward idea is to define all paired image and text instances to be relevant and all others irrelevant. However, this strategy produces a very small positive set and a much larger negative set, which often causes data imbalance during the training stage. A generalized yet more effective solution is to define a similarity matrix $\mathbf{S}\in\left\{0,1\right\}^{N\times N}$, so that when $S_{i,j}=1$ defines a positive pair $\left(i,j\right)$ and vice versa. The original sampling strategy is equivalent to $\mathbf{S}\equiv\mathbf{I}$.

Given $\mathbf{S}$, the objective of training involves minimizing the total distance with respect to $\boldsymbol{\theta}^\mathrm{I}$ and $\boldsymbol{\theta}^\mathrm{T}$, \textit{i.e.},
\begin{equation}\label{object function}
\boldsymbol{\theta}^{\mathrm{I},\star},\boldsymbol{\theta}^{\mathrm{T},\star}=\arg\min_{\boldsymbol{\theta}^\mathrm{I},\boldsymbol{\theta}^\mathrm{T}}=\sum_{i,j}S_{i,j}\cdot\left|\mathbf{f}_i^\mathrm{I}-\mathbf{f}_j^\mathrm{T}\right|.
\end{equation}
Therefore, the definition of $\mathbf{S}$ forms the major challenge of the learning task. According to whether extra labels of images and texts, besides the paired information, are used, existing methods can be categorized into either supervised or unsupervised learning. In the supervised setting, instance-level annotations (\textit{e.g.}, classification tags) are used to measure whether two instances are relevant, while in the unsupervised setting, no additional labels are available and thus the raw features are the only source of judgment. Obviously, the former provides more accurate estimation on $\mathbf{S}$ than the latter and, consequently, stronger models for cross-modal hashing. However, collecting additional annotations, even at the instance level, can be a large burden especially when the dataset is very large. Hence, we focus on improving the performance of unsupervised learning methods which are easier to be deployed to real-world scenarios.

%Additionly, we introduce the attempt of self-distillation. 
% discribe our teacher and student models and show how to distill knowledge from the teacher model for guiding the student model. We also analyse the effectiveness of the knowledge transferred from an unsupervised cross-modal hashing model, and explain the reason why it can be used to guide a supervised model. Additionly, we introduce our attempt of self-distillation. 
%-------------------------------------------------------------------------

\begin{figure}[t]
\begin{center}
\includegraphics[width=0.9\linewidth]{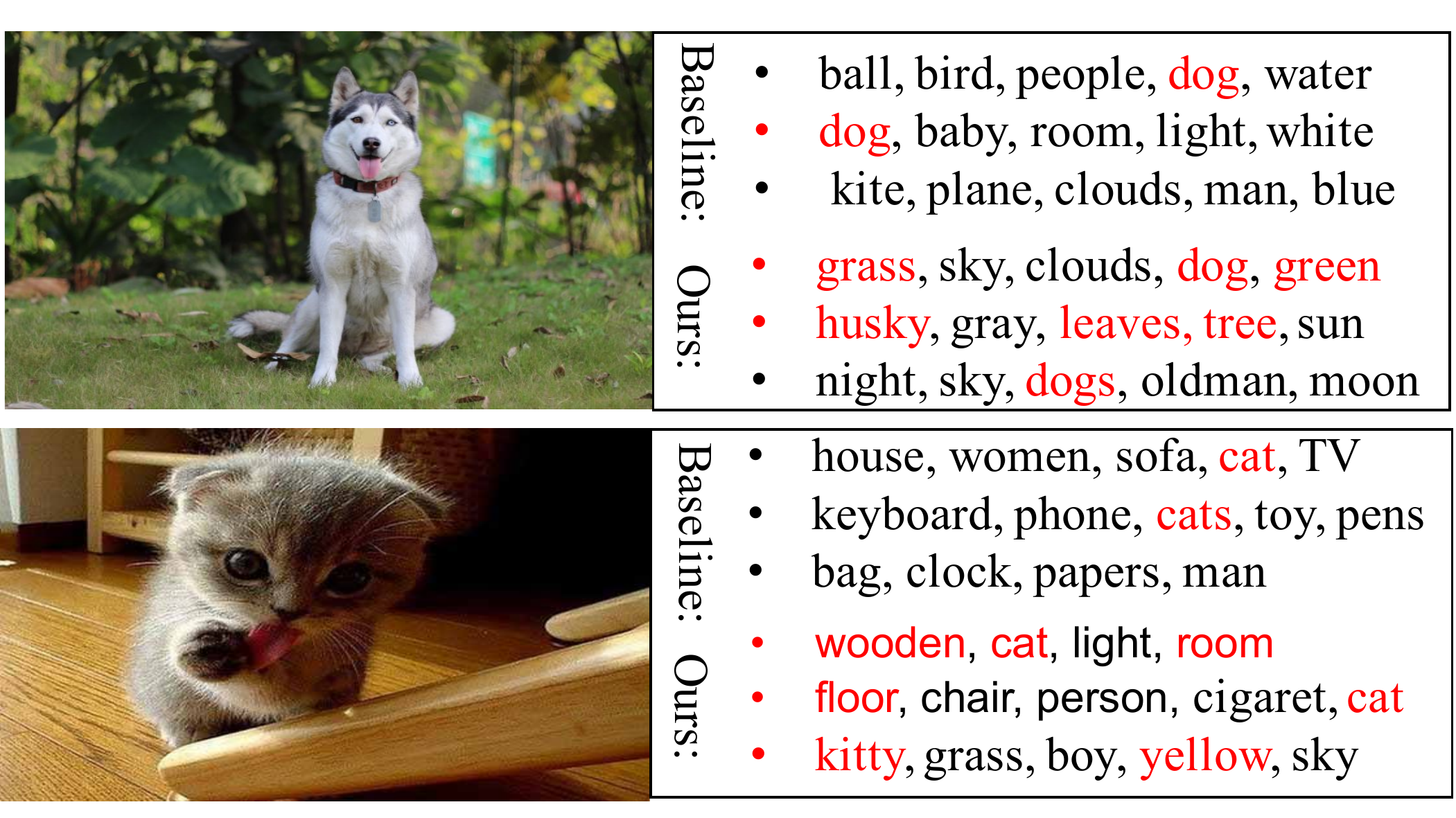}
\end{center}
\caption{Knowledge distilled from an unsupervised model (best viewed in color). Compared to the retrieval results in the original feature space, our approach produces more accurate information about the tag of an image and, more importantly, better estimation on the relevance of image-text pairs.}
\label{fig:similarity}
%\label{fig:onecol}
\end{figure}

\subsection{Unsupervised Knowledge Distillation}

\newcommand{\colwidthA}{1.0cm}
\begin{table*}[!t]
\centering
\begin{tabular}{|l||C{\colwidthA}|C{\colwidthA}|C{\colwidthA}||r|r|}
\hline
Function & new & image & text & P@$1000$ & P@$5000$ \\
\hline\hline
$S_{i,j}=\left(2-\left|\mathbf{v}_i^\mathrm{I}-\mathbf{v}_j^\mathrm{I}\right|_2\right)/2$
    &            & \checkmark &            & $74.6\%$ & $64.4\%$ \\
\hline
$S_{i,j}=\left(2-\left|\mathbf{v}_i^\mathrm{T}-\mathbf{v}_j^\mathrm{T}\right|_2\right)/2$
    &            &            & \checkmark & $57.1\%$ & $55.6\%$ \\
\hline
$S_{i,j}=\left(2-\left|\mathbf{f}_i^\mathrm{I}-\mathbf{f}_j^\mathrm{I}\right|_2\right)/2$
    & \checkmark & \checkmark &            & $84.6\%$ & $74.6\%$ \\
\hline
$S_{i,j}=\left(2-\left|\mathbf{f}_i^\mathrm{T}-\mathbf{f}_j^\mathrm{T}\right|_2\right)/2$
    & \checkmark &            & \checkmark & $75.9\%$ & $67.9\%$ \\
\hline
$S_{i,j}=\left(4-\left|\mathbf{f}_i^\mathrm{I}-\mathbf{f}_j^\mathrm{I}\right|_2-\left|\mathbf{f}_i^\mathrm{T}-\mathbf{f}_j^\mathrm{T}\right|_2\right)/4$
    & \checkmark & \checkmark & \checkmark & $83.9\%$ & $73.4\%$ \\
%\hline
%$S_{i,j}=\left(4-\left|\mathbf{f}_i^\mathrm{I}-\mathbf{f}_j^\mathrm{T}\right|_2-\left|\mathbf{f}_i^\mathrm{T}-\mathbf{f}_j^\mathrm{I}\right|_2\right)/4$
%    & \checkmark & \checkmark & \checkmark & \checkmark & $73.5\%$ \\
%$S_{i,j}=\max\left\{\left(2-\left|\mathbf{f}_i^\mathrm{I}-\mathbf{f}_j^\mathrm{T}\right|_2\right)/2,\left(2-\left|\mathbf{f}_i^\mathrm{T}-\mathbf{f}_j^\mathrm{I}\right|_2\right)/2\right\}$
%    & \checkmark & \checkmark & \checkmark & \checkmark & $??.?\%$ & \\
\hline
\end{tabular}
\caption{Comparison among different functions to measure the similarity between image-text pairs. All the results are computed using features extracted from a UGACH~\cite{zhang2018unsupervised} model trained on the MIRFlickr dataset. Here we consider four properties: `new' means that the new feature space, learned by the teacher model, is used; `image' and `text' means the corresponding features used, and `indiv' means image and text features are used individually. P@$1000$/P@$5000$ indicate the accuracy rates among the top $1000$/$5000$ retrieved pairs.}
\label{tab:functions}
\end{table*}

Our idea originates from the fact that, as shown above, the difference between supervised and unsupervised cross-modal hashing algorithms is not big, but supervised methods often report much higher accuracy than the unsupervised counterparts. Moreover, the supervised algorithms do not require \textit{real} supervision, namely, the manually labeled image/text tags, but only need to know, or estimate, the similarity between any pair of data, \textit{i.e.}, elements in $\mathbf{S}$. Beyond the unsupervised baseline that estimates $\mathbf{S}$ using raw image/text features (extracted from a pre-trained deep network or computed using bag-of-words statistics), we seek for the possibility that a cross-modal retrieval model, after trained in an unsupervised manner, can produce a more accurate estimation of $\mathbf{S}$. We illustrate an example in Figure~\ref{fig:similarity}. Later, we will show in experiments, with the help of \textit{oracle} annotations, that the updated estimation of $\mathbf{S}$ is indeed more accurate in terms of finding relevant pairs.

Note that the estimated $\mathbf{S}$ can be used to train either supervised or unsupervised models, with the formulations detailed above. When $\mathbf{S}$ is used for unsupervised learning, the only effect is to provide a better sampling strategy, so as to increase the portion of \textit{true-positive} image-text pairs in the chosen training set. This alleviates the risk that the model learns to pull the features of actually irrelevant pairs. When it is used for supervised learning, we are actually \textit{creating something from nothing}, \textit{i.e.}, guiding a supervised model with the output of an unsupervised model.

The proposed framework, \textbf{unsupervised knowledge distillation} (UKD), works as follows. After the teacher model has been trained, we obtain both $f^\mathrm{I}\!\left(\cdot;\boldsymbol{\theta}^\mathrm{I}\right)$ and $f^\mathrm{T}\!\left(\cdot;\boldsymbol{\theta}^\mathrm{T}\right)$ for image and text feature embedding, respectively. It remains to determine each element of $\mathbf{S}$. Without loss of generality, we assume that the feature vectors extracted from either modality, \textit{i.e.}, $\mathbf{f}_n^\mathrm{I}$ or $\mathbf{f}_i^\mathrm{T}$, have a $\ell_2$-norm. This is to ease the following calculations.

First, we point out that $S_{i,i}\equiv1$ for all $i$. When $i\neq j$, $S_{i,j}$ takes four vectors, $\mathbf{f}_i^\mathrm{I}$, $\mathbf{f}_i^\mathrm{T}$, $\mathbf{f}_j^\mathrm{I}$ and $\mathbf{f}_j^\mathrm{T}$, into consideration. The design of $S_{i,j}$ can have various forms. For example, it can consider both image and text features so that $S_{i,j}=\left(4-\left|\mathbf{f}_i^\mathrm{I}-\mathbf{f}_j^\mathrm{I}\right|_2-\left|\mathbf{f}_i^\mathrm{T}-\mathbf{f}_j^\mathrm{T}\right|_2\right)/4$, where $\left|\mathbf{f}_1-\mathbf{f}_2\right|_2$ is the Euclidean distance between two vectors which lies in the range of $\left[0,2\right]$ for two normalized vectors. Also, it is possible for $S_{i,j}$ to consider only single-modal information, \textit{e.g.}, $S_{i,j}=\left(2-\left|\mathbf{f}_i^\mathrm{I}-\mathbf{f}_j^\mathrm{I}\right|_2\right)/2$ in which only image features are used for measuring similarity.

Here, we take several definitions of $S_{i,j}$ into consideration, and compare their performance in finding true-positive pairs. Results are shown in Table~\ref{tab:functions}. We can observe several important properties that are useful for similarity measurement. \textbf{First}, the features trained for cross-modal hashing are indeed better than those without being fine-tuned; Second, measuring similarity in the image feature space is more accurate than that in the text feature space; Third, directly combining image and text similarity into one does not improve accuracy beyond using image similarity alone, though we expect that text features to provide auxiliary information. Motivated by these results, we use image features and text features to retrieve two lists of relevant pairs and then merge them into one. This strategy reports a precision of $76.1\%$ at top-$5000$ instances, surpassing that using image and text features alone. \textit{We fix this setting throughout the remainder of this paper.}

\subsection{Models and Implementation Details}

We first illustrate the supervised and unsupervised methods we have used. We take DCMH~\cite{jiang2017deep} as an example of supervised learning. Here we utilize the framework of DCMH but modify its architecture for higher accuracy. This model contains two deep neural networks, designed for the image modality and the text modality, respectively. The image modality network consists of $19$ layers, the first eighteen layers are the same as those in VGG19 network~\cite{simonyan2014very}, and the last layer maps features into the Hamming space. For the text modality, a multi-scale fusion model from SSAH~\cite{li2018self} which consists of multiple average pooling layers and a $1\times1$ convolutional layer is used to extract the text features. Then, a hash layer follows to map the text features into the Hamming space. 

On the other hand, we investigate UGACH~\cite{zhang2018unsupervised}, a representative unsupervised learning method as the teacher model. It consists of a generative module and a discriminative module. The discriminator receives the data selected by the generator as negative instances, and take the data sampled using $\mathbf{S}$ as positive instances. Then a triplet loss is used to optimize to obtain better discriminate ability for the discriminator. Both the generative and discriminative modules have a two-pathway architecture, each of which has two fully-connected layers. 
%The first layer maps the inputs into a common representation space, and the second layer maps the common representations into the Hamming space. 
We set the dimension of representation layer to $4096$ in our experiments. The dimension of the hash layer is same as the hash code length.

For the supervised model, we take the raw pixels as inputs. In pre-processing, we resize all images into $256\times256$ and crop a $224\times224$ patches randomly. We select relevant instances for the student model by using the teacher model with the highest precision ($128$ bits in all experiments). 
% \textcolor{red}{We chose $10\rm{,}000$ image-text pairs with smallest features to fed into the supervised student model, and $20$ for the unsupervised student model.  }
We set the number of relevant instances to be $10\rm{,}000$ for the supervised student model, and $20$ for the unsupervised student model. We train our approach in a batch-based manner and set the batch size to $256$. We train the model using an SGD optimizer with a weight decay of $0.01$. 
%The learning rate starts with $0.01$ and is decayed by a factor of $10$ after every two epochs. 
For the compared methods, we apply the same implementations as provided in the original work.

% \textcolor{red}{On the other hand, we investigates two unsupervised learning methods as the teacher model, namely, UGACH~\cite{zhang2018unsupervised} and UCH~\cite{li2019coupled}. Both these methods are composed of a generator and a discriminator. During training, the generator selects pairs of instances to challenge the discriminator, and the discriminator receives the generated pairs and the data sampled using $\mathbf{S}$, and tries to distinguish them FOR WHAT??????}

% We take DCMH~\cite{jiang2017deep} as an example of supervised learning. This model contains two deep neural networks, designed for the image modality and the text modality, respectively. In the learning process,

% Unlike the unsupervised scenario that both relevant and irrelevant pairs are constructed and then combine into the training process, only relevant pairs are used.%}

%-------------------------------------------------------------------------

\subsection{Relationship to Previous Work}

Our method is related to knowledge distillation~\cite{romero2014fitnets,zagoruyko2016paying,tarvainen2017mean}, which was proposed to extract knowledge from a teacher model to assist training a student model. Hinton~\textit{et al.}~\cite{hinton2015distilling} suggested that there should be some `dark knowledge' that can be propagated during this process. Recently, many efforts were made to study what the dark knowledge is~\cite{yim2017gift,yang2019training}, and/or how to efficiently take advantage of such knowledge~\cite{furlanello2018born,yu2019learning,wang2019progressive,cai2019exploring}. In particular, DarkRank~\cite{chen2018darkrank} distilled knowledge for deep metric learning by matching two probability distributions over ranking, while our approach utilized knowledge by selecting relevant instances. On the other hand, both~\cite{yu2019learning} and~\cite{park2019relational} transferred knowledge to improve the student models by designing a distillation loss, while our approach enables guiding a supervised method by an unsupervised method, in which no extra loss is used. 
%There are also works that used knowledge distillation improve unsupervised learning tasks in computer vision, \textit{e.g.}, monocular depth estimation~\cite{pilzer2019refine}.

We also notice the connection between our approach and the self-learning algorithms for semi-supervised learning, \textit{e.g.}, medical image analysis~\cite{zhou2019semi}. The shared idea is to start with a small part of labeled data (in our case, labeled image-text pairs) and try to explore the unlabeled part (in our case, other image-text pairs with unknown relevance), but the methods to gain additional supervision are different. Also, the idea that `training a stronger model at the second time' is related to the coarse-to-fine learning approaches~\cite{gu2018stack,zhou2017fixed} which often adopted iteration for larger improvements. 
%We also tried iteration and, from experimental results shown later, verify marginal gains in terms of retrieval performance.

Our approach shares the same idea with some prior work that guided a supervised model with the output of an unsupervised model. DeepCluster~\cite{caron2018deep} groups the features with a standard clustering algorithm and uses the subsequent assignments as supervision to update the weights of the network. Gomez~\textit{et al.}~\cite{gomez2017self} performed self-supervised learning of visual features by mining a large scale corpus of multi-modal (text and image) documents. Differently, our approach makes use of teacher-student optimization to combine the supervised and unsupervised models. Experiment results show the effectiveness of knowledge distillation.

%-------------------------------------------------------------------------

\section{Experiments}
\label{experiments}

%In this section, we test our approach for unsupervised cross-modal hashing on two benchmark datasets: MIRFlickr and NUS-WIDE. Firstly we introduce the datasets, baselines, evaluations and implementation details. Then we discribe the experiment results with detailed analusis.   
%-------------------------------------------------------------------------
\subsection{Datasets, Evaluation, and Baselines}

We evaluate our approach on two benchmark datasets: MIRFlickr and NUS-WIDE. MIRFlickr-25K~\cite{huiskes2008mir} consists of $25\rm{,}000$ images downloaded from Flickr. Each image is associated with text tags and annotated with at least one among $24$ pre-defined categories. Following UGACH~\cite{zhang2018unsupervised}, we use $20\rm{,}015$ image-text pairs in our experiments, where $2\rm{,}000$ are preserved as the query set and the rest are used for retrieving. We represent each image by a $4096$-dimensional feature vector, extracted from a pre-trained VGGNet~\cite{simonyan2014very} of $19$ layers, and each text by a $1386$-dimensional bag-of-words features.

NUS-WIDE~\cite{chua2009nus} is much larger than MIRFlickr, which contains $269\rm{,}498$ images and the associated text tags from Flickr. It defined $81$ categories, but there are considerable overlaps among them. Still, following UGACH~\cite{zhang2018unsupervised}, $10$ largest categories and the corresponding $186\rm{,}577$ image-text pairs are used in the experiments. We preserve $1\%$ of data as the query database and use the rest as the retrieval set. Each image is represented by a $4096$-dimensional feature vector extracted from the same VGGNet, and each text by a $1000$-dimensional bag-of-words vector.

Following the convention, we adopt the mean Average Precision (mAP) criterion to evaluate the retrieval performance of all methods. The mAP score is computed as the mean value of the average precision scores for all queries.

%-------------------------------------------------------------------------

\begin{table*}[!t]
\begin{center}
\setlength{\tabcolsep}{3.5mm} {
\begin{tabular}{|c|l||c|c|c|c||c|c|c|c|}
\hline
\multirow{2}{*}{Task}&\multirow{2}{*}{Method}&\multicolumn{4}{c||}{MIRFlickr-25K}&\multicolumn{4}{c|}{NUS-WIDE}\cr\cline{3-10}
    & &16&32&64&128&16&32&64&128\cr
\hline\hline
\multirow{11}{*}{$\mathrm{image}\rightarrow\mathrm{text}$}
 &CMSSH~\cite{bronstein2010data} &0.611&0.602&0.599&0.591&0.512&0.470&0.479&0.466 \\
 &SCM~\cite{zhang2014large} &0.636&0.640&0.641&0.643&0.517&0.514&0.518&0.518 \\
 &DCMH~\cite{jiang2017deep} &0.677&0.703&0.725&-&0.590&0.603&0.609&- \\
 &SSAH~\cite{li2018self} &0.797&0.809&0.810&-&0.636&0.636&0.637&- \\
 \cline{2-10}
 &CVH~\cite{kumar2011learning} &0.602&0.587&0.578&0.572&0.458&0.432&0.410&0.392 \\
 &PDH~\cite{rastegari2013predictable} &0.623&0.624&0.621&0.626&0.475&0.484&0.480&0.490 \\
 &CMFH~\cite{ding2016large} &0.659&0.660&0.663&0.653&0.517&0.550&0.547&0.520 \\
 &CCQ~\cite{long2016composite} &0.637&0.639&0.639&0.638&0.504&0.505&0.506&0.505 \\
 &UGACH~\cite{zhang2018unsupervised} &0.676&0.693&0.702&0.706&0.597&0.615&0.627&0.638 \\
 &UKD-US &0.695&0.703&0.705&0.707&0.606&0.621&0.634&0.643 \\
 &UKD-SS &\textbf{0.714}&\textbf{0.718}&\textbf{0.725}&\textbf{0.720}&\textbf{0.614}&\textbf{0.637}&\textbf{0.638}&\textbf{0.645} \\
\hline\hline
\multirow{11}{*}{$\mathrm{text}\rightarrow\mathrm{image}$}
 &CMSSH~\cite{bronstein2010data} &0.612&0.604&0.592&0.585&0.519&0.498&0.456&0.488 \\
 &SCM~\cite{zhang2014large} &0.661&0.664&0.668&0.670&0.518&0.510&0.517&0.518 \\
 &DCMH~\cite{jiang2017deep} &0.705&0.707&0.724&-&0.620&0.634&0.643&- \\
 &SSAH~\cite{li2018self} &0.782&0.797&0.799&-&0.653&0.676&0.683&- \\
 \cline{2-10}
 &CVH~\cite{kumar2011learning} &0.607&0.591&0.581&0.574&0.474&0.445&0.419&0.398 \\
 &PDH~\cite{rastegari2013predictable} &0.627&0.628&0.628&0.629&0.489&0.512&0.507&0.517 \\
 &CMFH~\cite{ding2016large} &0.611&0.606&0.575&0.563&0.439&0.416&0.377&0.349 \\
 &CCQ~\cite{long2016composite} &0.628&0.628&0.622&0.618&0.499&0.496&0.492&0.488 \\
 &UGACH~\cite{zhang2018unsupervised} &0.676&0.692&0.703&0.707&0.602&0.610&0.628&0.637 \\
 &UKD-US &0.704&0.707&0.715&0.714&0.621&0.625&0.640&0.647 \\
 &UKD-SS &\textbf{0.715}&\textbf{0.716}&\textbf{0.721}&\textbf{0.719}&\textbf{0.630}&\textbf{0.656}&\textbf{0.657}&\textbf{0.663} \\
\hline
\end{tabular}}
\end{center}
\caption{The mAP scores of our approach and state-of-the-art competitors, in two datasets and four different code lengths. In each half, the four rows above the horizontal line contain supervised learning algorithms, while the right rows below contain unsupervised ones.}
\label{Main table}
\end{table*}

We compare our approach against $9$ previous methods. $4$ of them used additional supervision (CMSSH~\cite{bronstein2010data}, SCM~\cite{zhang2014large}, DCMH~\cite{jiang2017deep}, and SSAH~\cite{li2018self}), and while $5$ others (CVH~\cite{kumar2011learning}, PDH~\cite{rastegari2013predictable}, CMFH~\cite{ding2016large}, and CCQ~\cite{long2016composite}), and UGACH~\cite{zhang2018unsupervised}) did not. Following our direct baseline, UGACH, we use a $19$-layer VGGNet~\cite{simonyan2014very} pre-trained on the ImageNet dataset~\cite{russakovsky2015imagenet} to extract deep features and, for a fair comparison, use them to replace the features used in other baselines, including those using handcrafted features.

%-------------------------------------------------------------------------
\subsection{Unsupervised Student vs. Supervised Student}

In Table~\ref{Main table}, we list the accuracy, in terms of mAP, of our approach as well as other methods for comparison. On two benchmark datasets MIRFlickr and NUS-WIDE. We use `$\mathrm{image}\rightarrow\mathrm{text}$' to denote the task that images are taken as the query to retrieval the instances in the text database, and `$\mathrm{text}\rightarrow\mathrm{image}$' the task in the opposite direction. Our approach is denoted by `UKD-US' and `UKD-SS', with `US' and `SS' indicating `unsupervised-student' and `supervised-student', respectively.

We observe interesting results. Regarding the $\mathrm{image}\rightarrow\mathrm{text}$ task, UKD-SS outperforms UKD-US significantly on the MIRFlickr dataset, but the advantage on the NUS-WIDE dataset becomes much smaller. This is explained by noting that the impact brought by supervision is different between these two datasets. We consider SSAH~\cite{li2018self} and UGACH~\cite{zhang2018unsupervised}, the supervised and unsupervised models we used as the students. SSAH typically outperforms UGACH by $9\%$ on MIRFlickr, but the number is quickly shrunk to $1\%$--$4\%$ on NUS-WIDE. This is partially due to the larger variance of the images in NUS-WIDE, which makes it difficulty for the labeled tags to provide accurate and valuable supervision. From this perspective, the reduced advantage of UKD-SS over UKD-US is reasonable, considering that SSAH is the upper-bound of UKD-SS.

On the other hand, by introducing extra supervision, (in particular, by checking the distance between the features extracted from an unsupervised model), considerable noise (\textit{e.g.}, inaccurate similarity measurement) is also introduced to the supervised student model. Hence, there is a tradeoff between the quality and impact of these self-annotated pairs. Most often, the latter can be measured by the advantage of the supervised student model over the unsupervised one, if both can be obtained in a small reference dataset.

%-------------------------------------------------------------------------
\subsection{Comparison to the State-of-the-Arts}

From Table~\ref{Main table}, one can observe that our approach, UKD, significantly outperforms all existing unsupervised cross-modal hashing methods on both datasets, and under any length of hash code. In particular, compared to our baseline (UGACH, which is also the strongest model that ever reported results with VGGNet-19 features), UKD enjoys $3.9\%$, $2.5\%$, $2.1\%$ and $1.3\%$ gains (averaged over $\mathrm{image}\rightarrow\mathrm{text}$ and $\mathrm{text}\rightarrow\mathrm{image}$) under $16$, $32$, $64$ and $128$ bits on the MIRFlickr dataset, and the corresponding numbers on the NUS-WIDE dataset are $2.3\%$, $3.4\%$, $2.0\%$ and $1.7\%$, respectively. Given such a high baseline, these improvements clearly demonstrate the effectiveness of distilling knowledge from the teacher model, although it is trained in an unsupervised manner. Moreover, the accuracy gain in more significant in the low-bit scenarios, arguably because richer information is provided by the teacher model which has $128$ bits. On the other hand, the amount of supervision saturates with the increasing number of compressed bits. We also tried to use full-precision models to serve as the teacher, but achieved marginal gain.

%-------------------------------------------------------------------------

\subsection{Does Iteration Help?}

Motivated by the consistent improvement from the teacher to the student, a question is straightforward: is it possible to further improve the performance if we continue distilling knowledge from the student, so as to guide a `new student'? We investigate this option, and results are summarized in Table~\ref{train in generations}. We find that, compared to the significant gain brought by the first knowledge distillation, the gain of the second round is mostly marginal, \textit{e.g.}, the average gain on the $\mathrm{image}\rightarrow\mathrm{text}$ task is $0.33\%$ compared to $0.60\%$ of the first round.

We owe this to the limited improvement of our student model in intra-modal learning -- recall that we have used intra-modal similarities to choose relevant pairs. Unlike the accuracy of cross-modal retreival performance, that of intra-model retrieval, from the teacher to the student, is hardly improved. This is to say, the new batch of image-text pairs for either supervised or unsupervised learning do not have a clear advantage over the previous batch, and so the quality of training data mostly remains unchanged.

\begin{table}[!t]
%\label{train in generations}
\begin{center}
\setlength{\tabcolsep}{1.5mm} {
\begin{tabular}{|c|c|c|c|c|c|c|c|c|}
\hline
\multirow{2}{*}{Task}&\multirow{2}{*}{Method}&\multicolumn{4}{c|}{MIRFlickr-25K}\cr\cline{3-6}
    & &16&32&64&128\cr
\hline   
\multirow{3}{*}{$\mathrm{image}\rightarrow\mathrm{text}$}
 &GEN-0 &0.676&0.693&0.702&0.706 \\
 &GEN-1 &0.695&0.703&0.705&0.707 \\
 &GEN-2 &0.698&0.705&0.708&0.712 \\
\hline   
\multirow{3}{*}{$\mathrm{text}\rightarrow\mathrm{image}$}
 &GEN-0 &0.676&0.692&0.703&0.707 \\
 &GEN-1 &0.704&0.707&0.715&0.714 \\
 &GEN-2 &0.705&0.712&0.716&0.719 \\
\hline   
\end{tabular} }
\end{center}
\caption{Results of training in generations for unsupervised student model on MIRFlickr-25K. `GEN-0' and `GEN-1' are identical to the UGACH and UKD-US models reported in Table~\ref{Main table}, respectively.}
\label{train in generations}
\end{table}

\begin{table}[!t]
\begin{center}
\setlength{\tabcolsep}{1.5mm} {
\begin{tabular}{|c|c|c|c|c|c|}
\hline
\multirow{2}{*}{Method} & \multirow{2}{*}{Task} & \multicolumn{4}{c|}{MIRFlickr-25K} \\ \cline{3-6} 
& & 16 & 32 & 64 & 128 \\ \hline
\multirow{2}{*}{UKD-SS} 
& $\mathrm{image}\rightarrow\mathrm{text}$            
& 0.711 & 0.704 & 0.711 & 0.720 \\ \cline{2-6} 
& $\mathrm{text}\rightarrow\mathrm{image}$  
& 0.692  & 0.702 & 0.705 & 0.706 \\ \hline
\end{tabular} }
\end{center}
\caption{Results of using a 16-bit teacher to guide the supervised student model on MIRFlickr-25K.}
\label{Small bits teacher}
\end{table}

\begin{table}[!t]
\begin{center}
\setlength{\tabcolsep}{1.5mm} {
\begin{tabular}{|c|l||c|c|c|}
\hline
\multirow{2}{*}{Task}&\multirow{2}{*}{Method}&\multicolumn{3}{c|}{MIRFlickr-25K}\cr\cline{3-5}
    & &16&32&64\cr
\hline\hline   
\multirow{4}{*}{$\mathrm{image}\rightarrow\mathrm{text}$}
 &UGACH~\cite{zhang2018unsupervised} &0.603&0.607&0.616 \\
 &UCH~\cite{li2019coupled} &0.654&0.669&0.679 \\
 &UKD-US &0.667&0.674&0.677 \\
 &UKD-SS &0.678&0.680&0.679 \\
\hline\hline
\multirow{4}{*}{$\mathrm{text}\rightarrow\mathrm{image}$}
 &UGACH~\cite{zhang2018unsupervised} &0.590&0.632&0.642 \\
 &UCH~\cite{li2019coupled} &0.661&0.667&0.668 \\
 &UKD-US &0.676&0.683&0.680 \\
 &UKD-SS &0.688&0.687&0.694 \\
\hline 
\end{tabular} }
\end{center}
\caption{Accuracy (mAP) comparison on MIRFlickr-25K, with UGACH and UCH as the baselines. To observe how a stronger teacher model ($128$-bit) teaches a weaker student model, we only report $16$-bit, $32$-bit and $64$-bit results.}
\label{SOTA compare}
\end{table}

%-------------------------------------------------------------------------
\subsection{Diagnostic Experiments}

\noindent$\bullet$\quad\textbf{Knowledge Distillation with a Weaker Teacher}

In order to show that UKD can work under a relatively weaker teacher signal, we use a $16$-bit model of UGACH~\cite{zhang2018unsupervised} as the teacher. As shown in Table~\ref{Small bits teacher}, we still achieve consistent accuracy gain beyond the baseline. However, the gain is reduced compared to using a $128$-bit teacher, since the benefit of UKD is mostly determined by the quality of the similarity matrix $\mathbf{S}$, and a weaker teacher often leads to a weaker $\mathbf{S}$, \textit{e.g.}, the precision of the top-ranked list of pairs is reduced.

\vspace{0.2cm}
\noindent$\bullet$\quad\textbf{Transferring to Other Features}

To verify that our approach is generalized to other features, we apply it to UCH~\cite{li2019coupled}, a recently published unsupervised cross-modal hashing method, using features extracted from a pre-trained CNN-F model~\cite{chatfield2014return} (same as in the original paper). Table~\ref{SOTA compare} shows the comparison between UCH and our approach in terms of mAP values on MIRFlickr. Note that our baseline is still UGACH, with the features replaced, since the authors of UCH did not provide the code. One can see that both UKD-US and UKD-SS outperform UGACH (and also, UCH), and UKD-SS works better than UKD-US, \textit{i.e.}, the same phenomena we have observed previously.

%This is because UCH just utilize the existing pairwise information to learn modality correlation, which may cause a worse generalization to test data. By exploiting extensive relevance information from the outputs produced by an unsupervised teacher model, we can accurately capture modality correlation to improve the performance of cross-modal hashing. Additionally, "S-S" achieves better performance than the "U-S". It further verfies the effectiveness of utilizing an unsupervised method to guide a supervised method for cross-modal hashing. 

\begin{figure}[!t]
\centering    
%\begin{minipage}[t]{0.45 \linewidth}
%\centering          
\includegraphics[width=0.9\linewidth]{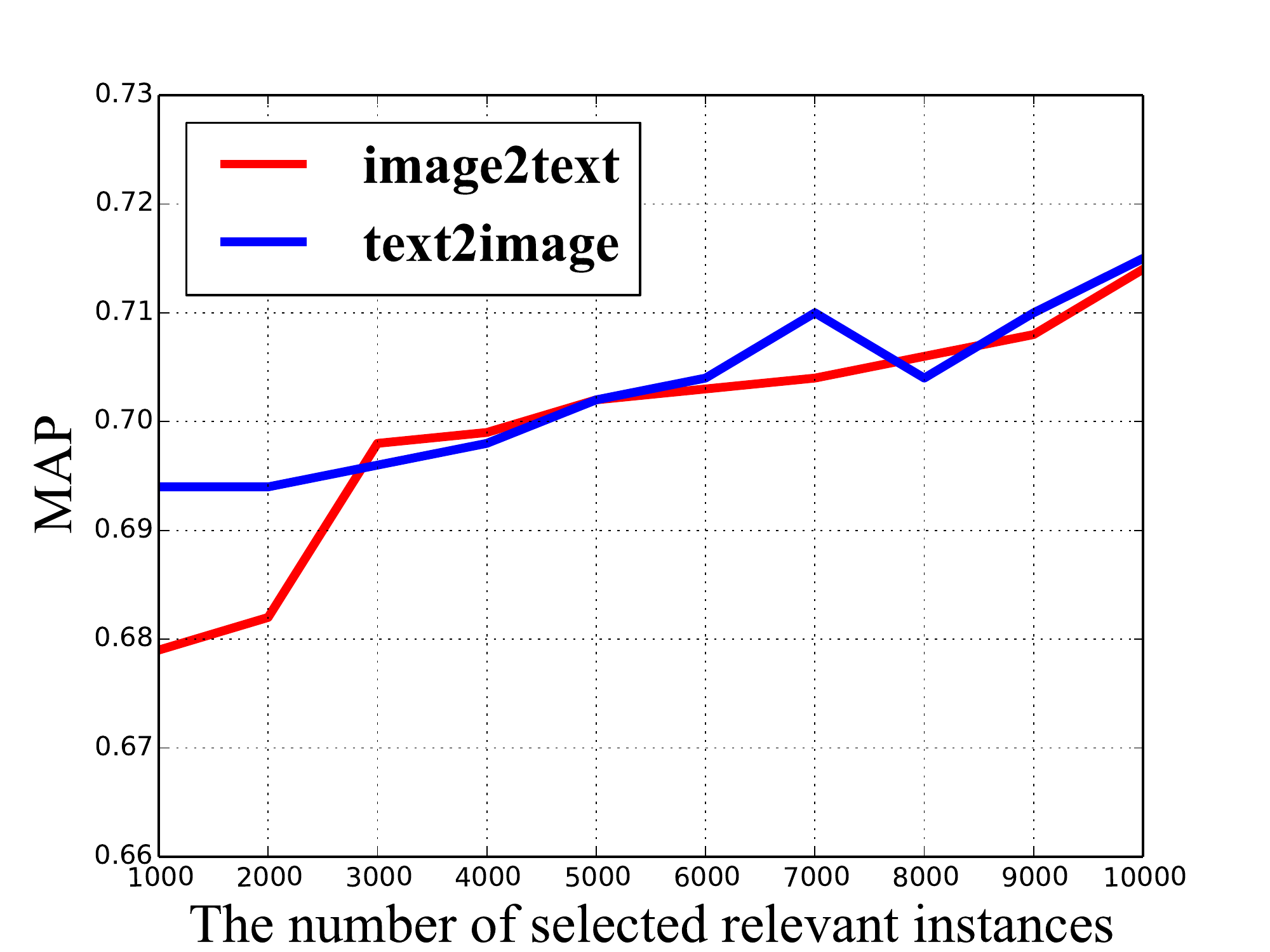}
%\end{minipage}
% \begin{minipage}[t]{0.45 \linewidth}
% \centering  
% \includegraphics[height=3.5cm, width=1.6in]{U-S.eps}
% \end{minipage}
\caption{The mAP value with respect to the number of relevant pairs selected (tested on the MIRFlickr dataset, teacher is a $128$-bit model, student is a $16$-bit model).} 
\label{fig:compare s-s with u-s}  
\end{figure}

\begin{figure}[!t]
\begin{center}
\includegraphics[width=0.9\linewidth]{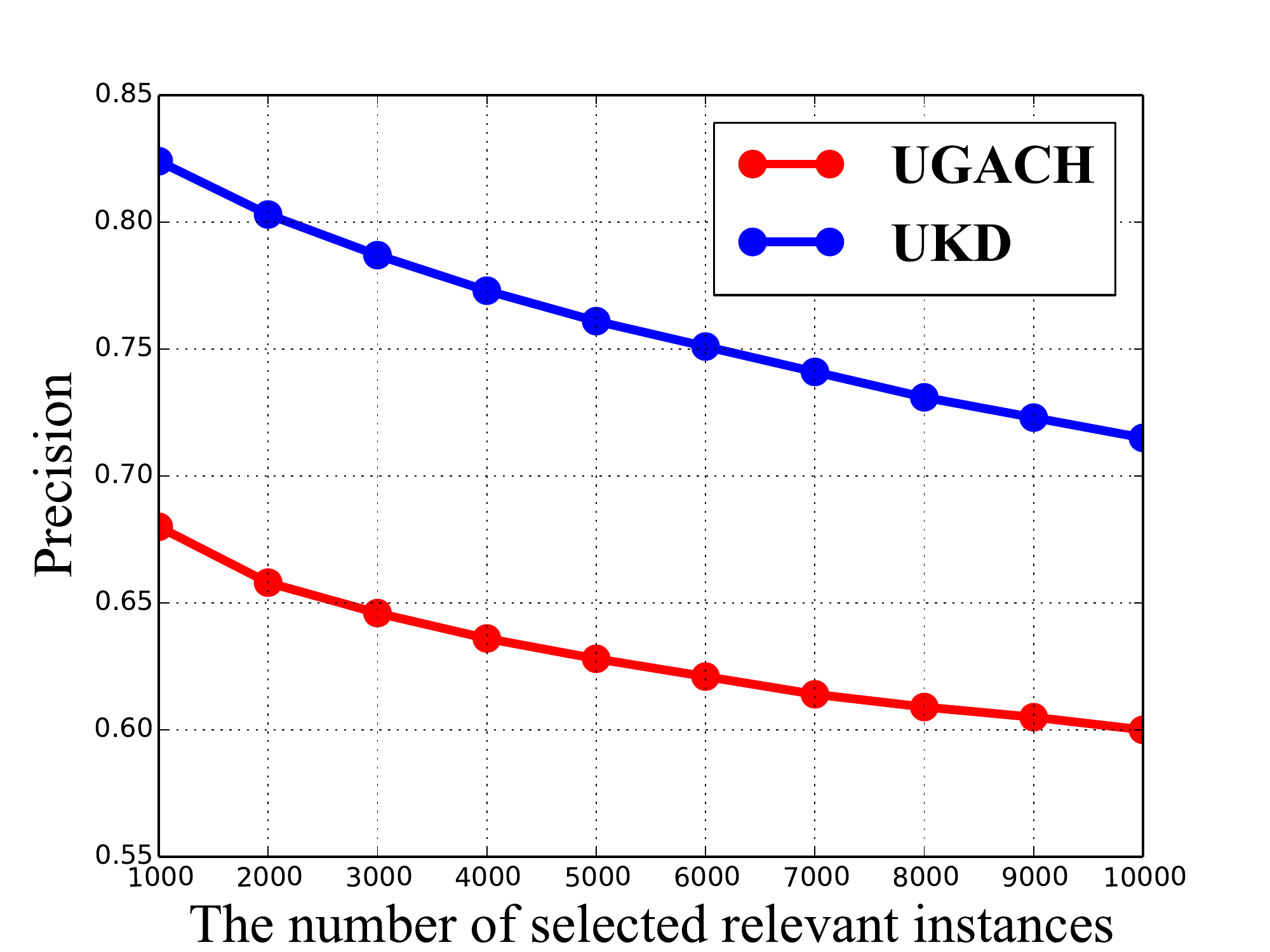}
\end{center}
\caption{The top-$K$ precision curves with respect to the number of relevant pairs selected (tested on the MIRFlickr dataset, teacher is a $128$-bit model).}
\label{fig:precision}
\end{figure}

\vspace{0.2cm}
\noindent$\bullet$\quad\textbf{Sensitivity to the Number of Selected Pairs}

%Next, we analyse the superiority of the supervised student model when compared to the unsupervised one. To achieve this, we investigate the the variation of performance for both of them when using different size of relevant instances. Figure~\ref{fig:compare s-s with u-s} shows the MAP values of image$\rightarrow$text/text$\rightarrow$image vary with the increase of $k$ on MIRFlickr when code length is 16. The left presents the results of supervised student model and the right shows the results of unsupervised student model. The $k$ denotes the number of relevant instances for each query selected by using the outputs of the teacher model. 

Next, we analyze how the performance of cross-modal hashing is related to the number of relevant pairs selected during the training process. In Figure~\ref{fig:compare s-s with u-s}, one can observe a trend of accuracy gain as the number of selected pairs increases, but when the number goes to a relatively large value, it tends to saturate and even goes down a little bit. This is related to the total number of relevant pairs in the dataset and, of course, the ability of the model in choosing relevant pairs.

We also compare our approach with the baseline in terms of the precision of the top-ranked, selected instance pairs. From Figure~\ref{fig:precision}, we can see that UKD enjoys a significant advantage over UGACH, our direct baseline. Nevertheless, we see a rapid drop in precision when the number of selected pairs grows, implying that non-top-ranked pairs can introduce noise to the model. Again, this is a tradeoff between quantity and quality.

\vspace{0.2cm}
\noindent$\bullet$\quad\textbf{Qualitative Studies}

% \begin{figure}[!t]
% \begin{center}
% \includegraphics[width=\linewidth]{selected_pairs.pdf}
% \end{center}
% \caption{Examples of relevant instances found by measuring the distance between the output features of the teacher model (UGACH) on MIRFlickr. Top: a text query with top-$3$ retrieved instances; bottom: an image query with top-$3$ retrieved instances.}
% \label{fig:similar instances}
% \end{figure}

%Figure~\ref{fig:similar instances} presents some examples of relevant instances selected by using the outputs of the teacher model. The top part shows several similar images from a text query and the bottom part shows similar texts from an image query. If we only consider top-ranked retrieval results, the instances are relatively accurate regardless whether the query is an image or a text. This again suggests that the effectiveness of the UKD approach comes from mining additional relevant data from the unlabeled pool of the dataset.

\begin{figure}[!t]
\begin{center}
\includegraphics[width=\linewidth]{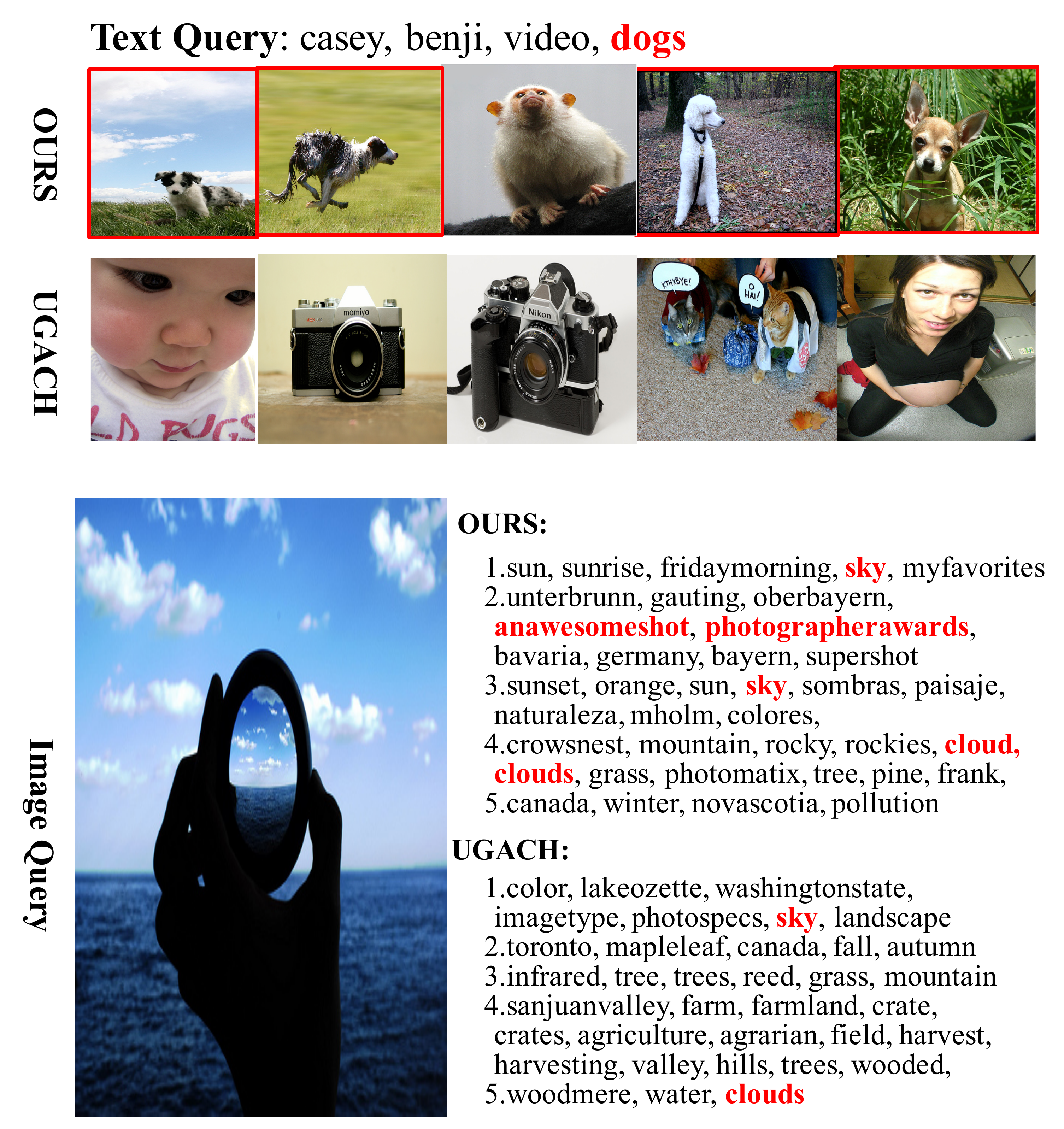}
\end{center}
\caption{Qualitative comparison (top: a text query with top-$5$ retrieved instances; bottom: an image query with top-$5$ retrieved instances) between our approach and UGACH ($16$-bit hashing), our direct baseline. Red frames and words indicate relevant images or words in the retrieved results. Note that the image query is much more difficult, as it contains semantically complicated concepts which even requires aesthetic perception to understand.} %(\textit{e.g.}, \textit{awesome-scene}, \textit{photographer-awards}) which even requires aesthetic perception to understand.}
\label{fig:visual results}
\end{figure}

Finally, we qualitatively compare the results of our approach and the baseline. Figure~\ref{fig:visual results} shows two typical examples. The $\mathrm{text}\rightarrow\mathrm{image}$ query (\textit{dog}) is relatively simple, but in the original paired training set, there are no sufficient amount of labeled data for the algorithm to learn the vision-language correspondence. This is compensated with the enlarged set found by an unsupervised teacher model. In comparison, the $\mathrm{image}\rightarrow\mathrm{text}$ query contains complicated semantics 
%(\textit{e.g.}, \textit{awesome-scene}, \textit{photographer-awards}) 
that are even more difficult to learn, but our model, by making use of image-level similarity, mines extra training data from other sources (see the examples in Figure~\ref{fig:visual results} which is also related to these tags). Consequently, the prediction of our approach is much better.

%-------------------------------------------------------------------------
\section{Conclusions}
\label{conclusions}

%\textcolor{blue}{
In this paper, we propose a novel approach to improve cross-modal hashing which enables guiding a supervised method using the outputs produced by an unsupervised method. We make use of teacher-student optimization for propagating knowledge. Superior performance can be achieved for the supervised student model by utilizing the extensive relevance information exploited from the outputs of the unsupervised teacher model. We evaluate our approach on two benchmarks MIRFlickr and NUS-WIDE, and the experiment results show that our method outperforms the state-of-the-art methods. 
%In the future, we will design a more powerful network architecture cooperates with the UKD framework. 

%In the future, we will investigate a method to utilize the teacher model to adaptively select relevant instances for each query, which is expected to achieve better performance.

%Unsupervised knowledge distillation also enjoys the advantage of being generalized. To realize this, we emphasize that many unsupervised cross-modal hashing methods (\textit{e.g.}, UGACH~\cite{zhang2018unsupervised} and UCH~\cite{li2019coupled}) have limitations on the network design because of lacking supervision information. Our approach, by exploiting relevance information from the outputs of an unsupervised model, has the potential of being more flexible on the student network design.

%This work leaves a lot of unsolved problems. For example, we are encountering the problem that the number of selected pairs largely impacts the performance of the student model. As we have clarified before, this is a tradeoff between the quantity and quality of supervision, but our approach lacks an effective criterion in judging if the balance has been arrived. These topics are left for future research.

%-------------------------------------------------------------------------
\vspace{0.2cm}
\textbf{Acknowledgements}\quad
This work was supported in part by the National Natural Science Foundation of China under grant 61722204, 61932009, and in part by the National Key Research and Development Program of China under grant 2019YFA0706200, 2018AAA0102002.

%-------------------------------------------------------------------------

\bibliographystyle{ieee_fullname}
\bibliography{egpaper_final}

\end{document}